\documentclass[conference]{IEEEtran}
\IEEEoverridecommandlockouts
\usepackage{cite}
\usepackage{multirow}
\usepackage[pdftex]{graphicx}
\usepackage[cmex10]{amsmath}
\usepackage{amssymb}
\usepackage{caption}
\usepackage[breaklinks=true,bookmarks=false]{hyperref}
\usepackage{bm}
\usepackage[inline]{enumitem}
\usepackage{gensymb}
\usepackage{tabu}
\usepackage{pifont}

\newcommand{\etal}{\textit{et al}. }
\newcommand{\eg}{\emph{e.g}. }
\newcommand{\ie}{\emph{i.e}. }

\newcommand\T{\rule{0pt}{2.9ex}}       
\newcommand\B{\rule[-1.2ex]{0pt}{0pt}} 

\newcommand{\cmark}{\ding{51}}%
\newcommand{\xmark}{\ding{55}}%

\begin{document}

\title{\emph{A King's Ransom for Encryption}:\\Ransomware Classification using Augmented One-Shot Learning and Bayesian Approximation}

\author{
\IEEEauthorblockN{Amir Atapour-Abarghouei,
Stephen Bonner and
Andrew Stephen McGough}

\IEEEauthorblockA{ School of Computing, Newcastle University, Newcastle, UK  \\ \{amir.atapour-abarghouei, stephen.bonner3, stephen.mcgough\}@newcastle.ac.uk} 

}

\maketitle

\begin{abstract}

    Newly emerging variants of ransomware pose an ever-growing threat to computer systems governing every aspect of modern life through the handling and analysis of big data. While various recent security-based approaches have focused on detecting and classifying ransomware at the network or system level, easy-to-use post-infection ransomware classification for the lay user has not been attempted before. In this paper, we investigate the possibility of classifying the ransomware a system is infected with simply based on a screenshot of the splash screen or the ransom note captured using a consumer camera commonly found in any modern mobile device. To train and evaluate our system, we create a sample dataset of the splash screens of 50 well-known ransomware variants. In our dataset, only a single training image is available per ransomware. Instead of creating a large training dataset of ransomware screenshots, we simulate screenshot capture conditions via carefully designed data augmentation techniques, enabling simple and efficient one-shot learning. Moreover, using model uncertainty obtained via Bayesian approximation, we ensure special input cases such as unrelated non-ransomware images and previously-unseen ransomware variants are correctly identified for special handling and not mis-classified. Extensive experimental evaluation demonstrates the efficacy of our work, with accuracy levels of up to 93.6\% for ransomware classification. 

\end{abstract}

\begin{IEEEkeywords}
Machine Learning, Ransomware Classification, Model Uncertainty, Bayesian Approximation, One-Shot Learning\vspace{-0.5cm}
\end{IEEEkeywords}

\section{Introduction}
\label{sec:introduction}

Due to the increasingly prominent role of the internet in various facets of modern life, any malicious online activity has the potential to disrupt the social order, sometimes with dire repercussions. Of the numerous variants of malware often spread for economic gain, ransomware has recently received significant attention within the cybersecurity community \cite{zhang2019classification} due to its wide range of targets, the significant harm it can inflict on the victims, the great financial incentive it provides for organised crime syndicates and its constant evolution, allowing its variants to regularly bypass state-of-the-art anti-virus and anti-malware \cite{kok2019ransomware}.

There are, in essence, two types of ransomware: locker ransomware, which locks the targeted system and prevents or constrains user access, but is often easily resolvable for a technically-savvy user, and crypto-ransomware, which can be significantly more difficult to deal with and can lead to irreversible harm as it encrypts files within the targeted system. A third type of ransomware, called scareware, attempts to scare lay users into paying the ransom without actually damaging the computer in any way \cite{al2018ransomware} only using an intimidating splash screen. This substantial level of diversity among ransomware variants gives significant importance to a robust classification system that could easily identify the ransomware and guide the victims towards appropriate support.

Various classification and detection techniques within the existing literature \cite{zhang2019classification, 8125850, ferrante2017extinguishing} facilitate identifying and countering ransomware attacks for technically-adept individuals and organisations with a large security and IT infrastructure. However, ransomware classification methods tailored towards the laypersons, which make up the majority of users and are often targeted easily, are scarce. In this paper, we propose an image classification pipeline, which enables any individual to identify the variant of the ransomware they are infected with based on a screenshot of the splash screen or the ransom note casually captured using a consumer-grade camera, such as those commonly found in any modern mobile phone.

\begin{figure}[t!]
	\centering
	\includegraphics[width=0.99\linewidth]{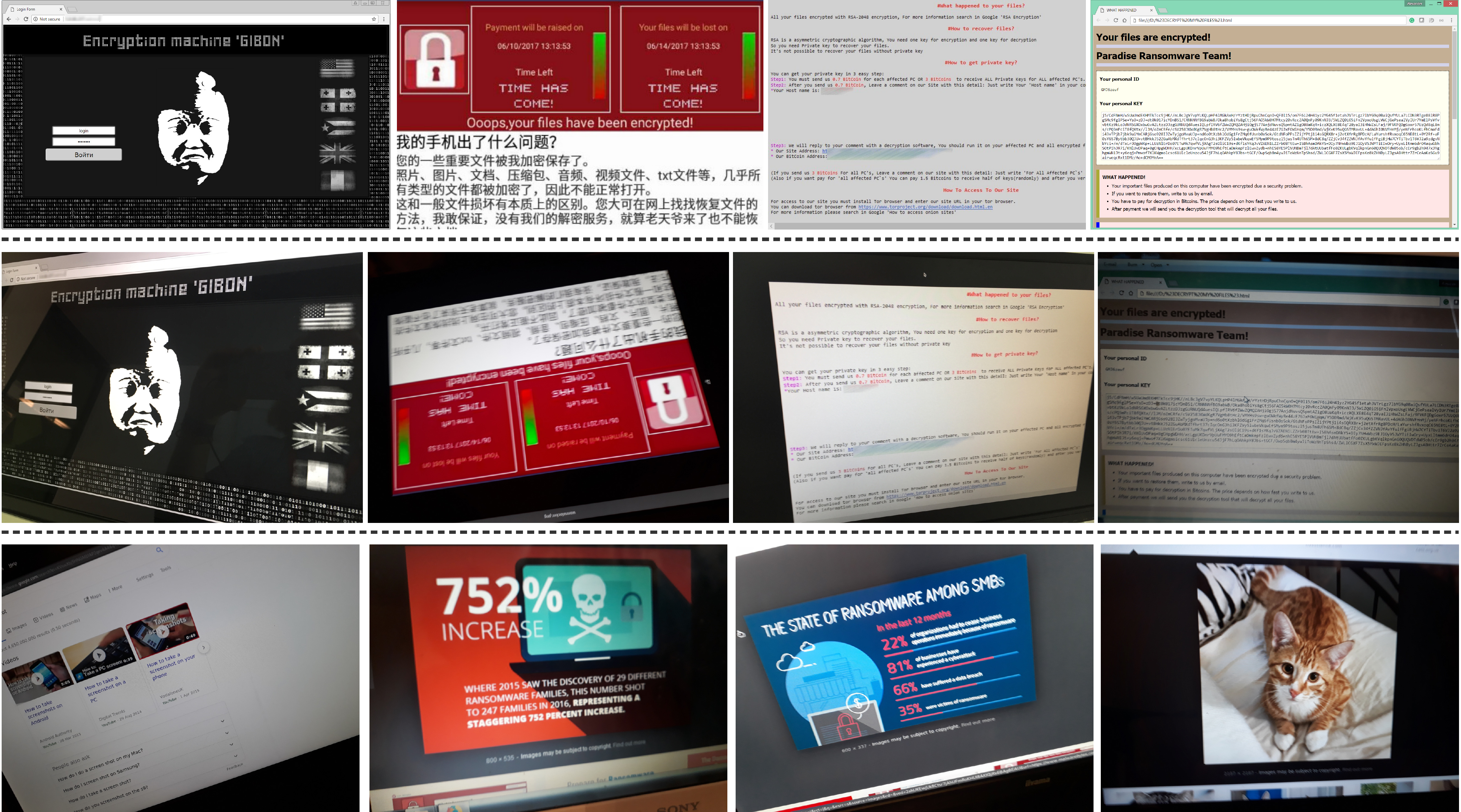}
	\captionsetup[figure]{skip=7pt}
	\captionof{figure}{Examples from our dataset: training images (ransomware splash screens - \emph{top}), positive test data (screenshots of ransomware splash screens - \emph{middle}) and negative test data (unrelated images for uncertainty evaluation - \emph{bottom}).}%
	\label{fig:taste_data}\vspace{-0.4cm}
\end{figure}

While a significant portion of the literature on classification has been dedicated to achieving consistent high-accuracy results using a variety of optimised deep neural networks \cite{simonyan2014very, he2016deep, iandola2016squeezenet, huang2017densely, szegedy2016rethinking, ma2018shufflenet, sandler2018mobilenetv2, xie2017aggregated}, most of these techniques require large quantities of accurately-labelled data, which for our task, translates to a large corpus of splash screen images captured from computer screens under different environmental conditions (lighting, field of view, camera angle, etc.) varied enough to simulate any future image capture and thus avoid over-fitting. A na\"ive solution to the data requirement problem would be to accept the considerable costs and resources required to create such a large dataset, but in this work, we attempt to circumvent the need for big data by recreating the conditions that lead to the appearance of a screenshot by means of carefully designed and tuned data augmentation techniques. In essence, our \emph{one-shot} learning framework is capable of classifying any image of a ransomware splash screen captured using a camera by only ever seeing a single original image for each class of ransomware. This enables our approach to rapidly learn to classify new variants if the model is simply retrained or fine-tuned using a single training image. Consequently, our dataset consists of a single image per variant of ransom note or splash screen for training and ten screenshots of said ransom notes captured using a mobile phone camera for testing (Figure \ref{fig:taste_data}).

\begin{figure}[t!]
	\centering
	\includegraphics[width=0.99\linewidth]{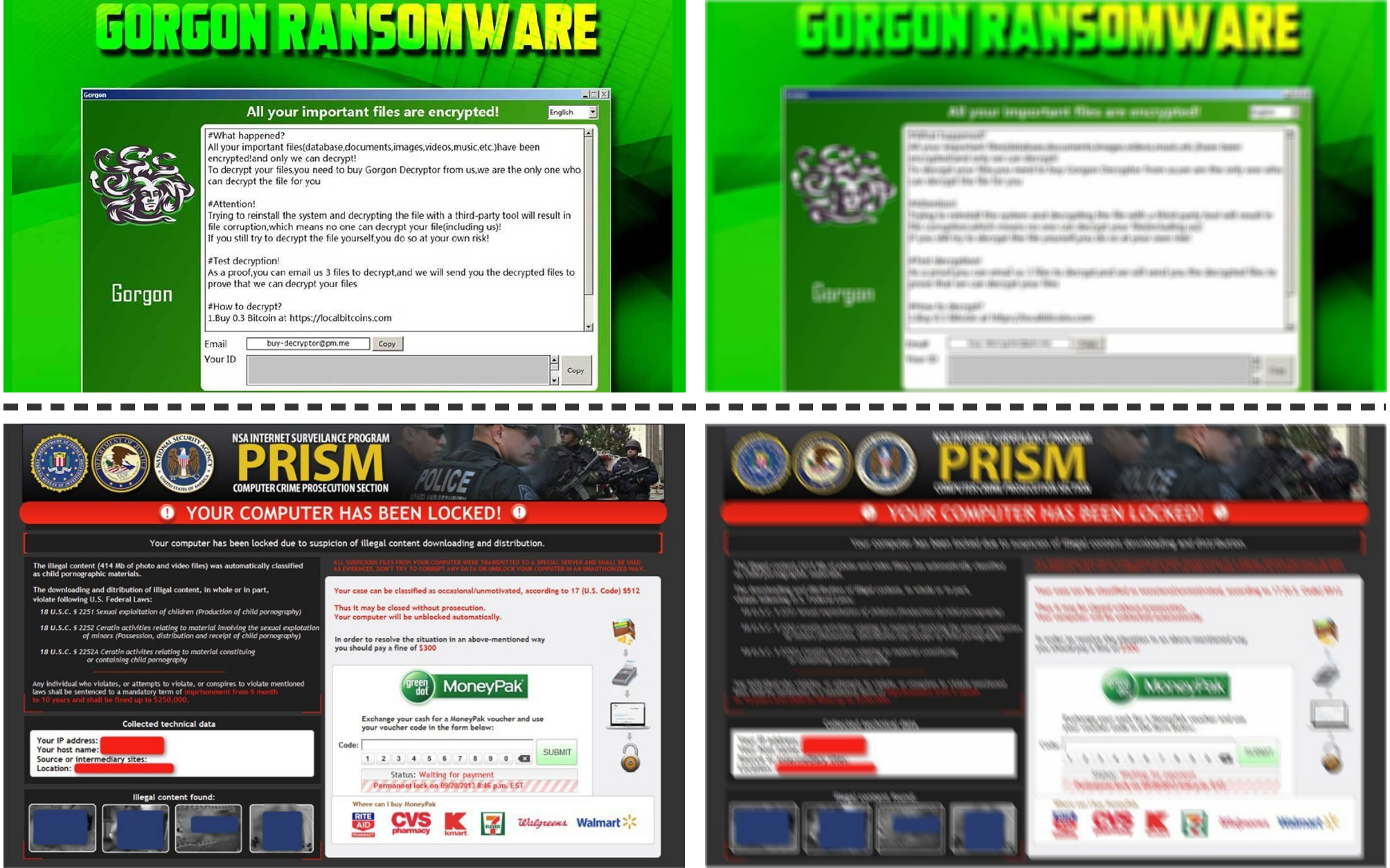}
	\captionsetup[figure]{skip=7pt}
	\captionof{figure}{Examples of augmentation methods used to simulate screenshot capture: defocus blur (top), motion blur (bottom).}%
	\label{fig:augs}\vspace{-0.6cm}
\end{figure}

Additionally, modern neural-based classification approaches are notoriously known for attempting to classify inputs on which they have not been trained \cite{Karmakar2018444} or completely miss-classifying images sampled from distributions with slight deviations from the training set \cite{kurakin2017adversarial}. This means an off-the-shelf approach will wrongly classify any unrelated input (\eg non-ransomware images, images of new ransomware variants unknown to the existing model, carefully-designed adversarial examples), sometimes with a high degree of confidence. To remedy this, we turn towards the recent advances in variational inference and its implications in calculating model uncertainty in neural networks \cite{gal2016dropout, gal2017concrete, kingma2015variational, hron2017variational, hron2018variational}. Not only does the integration of Bayesian inference into a neural network make it more robust against adversarial attacks, access to model uncertainty enables the network to reject irrelevant inputs sampled from outside the distribution of the training data. The inclusion of model uncertainty calculations in our pipeline requires its very own evaluation methodology, for which purpose, we also include a negative test set (Figure \ref{fig:taste_data} -- \emph{bottom}) in our dataset to assess our uncertainty values. This dataset consists of unrelated input images which the model should be uncertain about as it has not been trained to classify such images. In short, the primary contributions of this work are as follows:

\begin{itemize}
  \item \emph{Ransomware Classification:} We provide a simple pipeline that enables any laypersons to identify the variant of ransomware they have been infected with by casually taking a photograph of their computer screen displaying the ransom note or splash screen.
  
  \item \emph{One-Shot Learning through Data Augmentation:} We investigate the possibility of using different data augmentation techniques to simulate the appearance of a screenshot given the original splash screen, thereby enabling training on a single data point per class with significant generalisation capabilities.  
  
  \item \emph{Model Uncertainty via Bayesian Approximation:} We explore the use of various forms of Bayesian inference to further improve generalisation and obtain model uncertainty to avoid classifying unrelated images and as-of-yet-unknown variants of ransomware. 
\end{itemize}

To enable easier reproducibility, the source code, pre-trained models and the dataset will be publicly available.\footnote{\href{https://github.com/atapour/ransomware-classification}{https://github.com/atapour/ransomware-classification}}.

\begin{figure}[t!]
	\centering
	\includegraphics[width=0.99\linewidth]{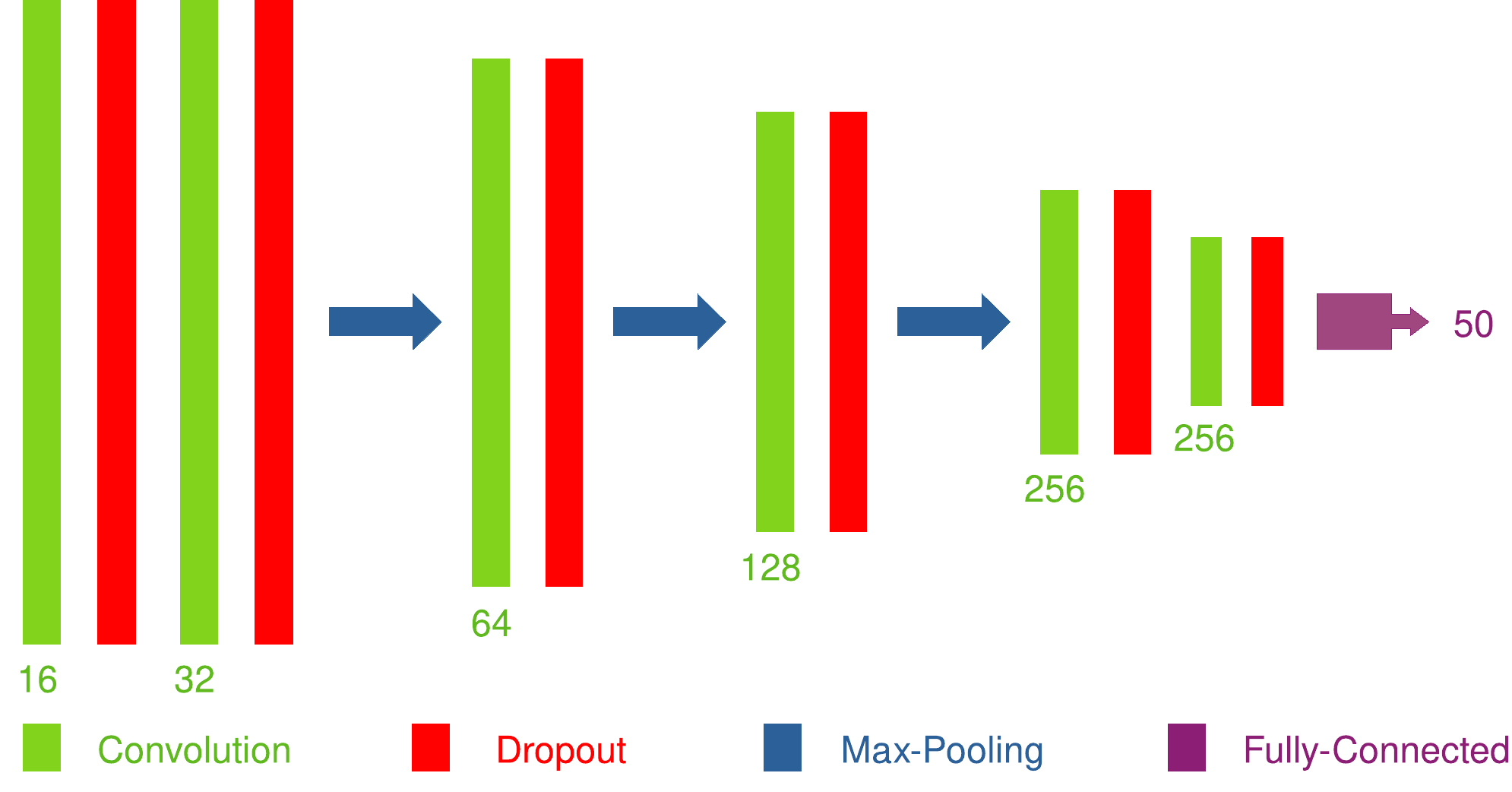}
	\captionsetup[figure]{skip=7pt}
	\captionof{figure}{The custom architecture used in our experiments.}%
	\label{fig:arch}\vspace{-0.3cm}
\end{figure}

\section{Related Work}
\label{sec:lit}

We consider relevant prior work over three distinct areas, ransomware classification and detection (Section \ref{ssec:lit:ransomware}), one-shot learning (Section \ref{ssec:lit:one_shot}), and Bayesian approximation (Section \ref{ssec:lit:bayesian}).

\begin{figure*}[t!]
	\centering
	\includegraphics[width=0.99\linewidth]{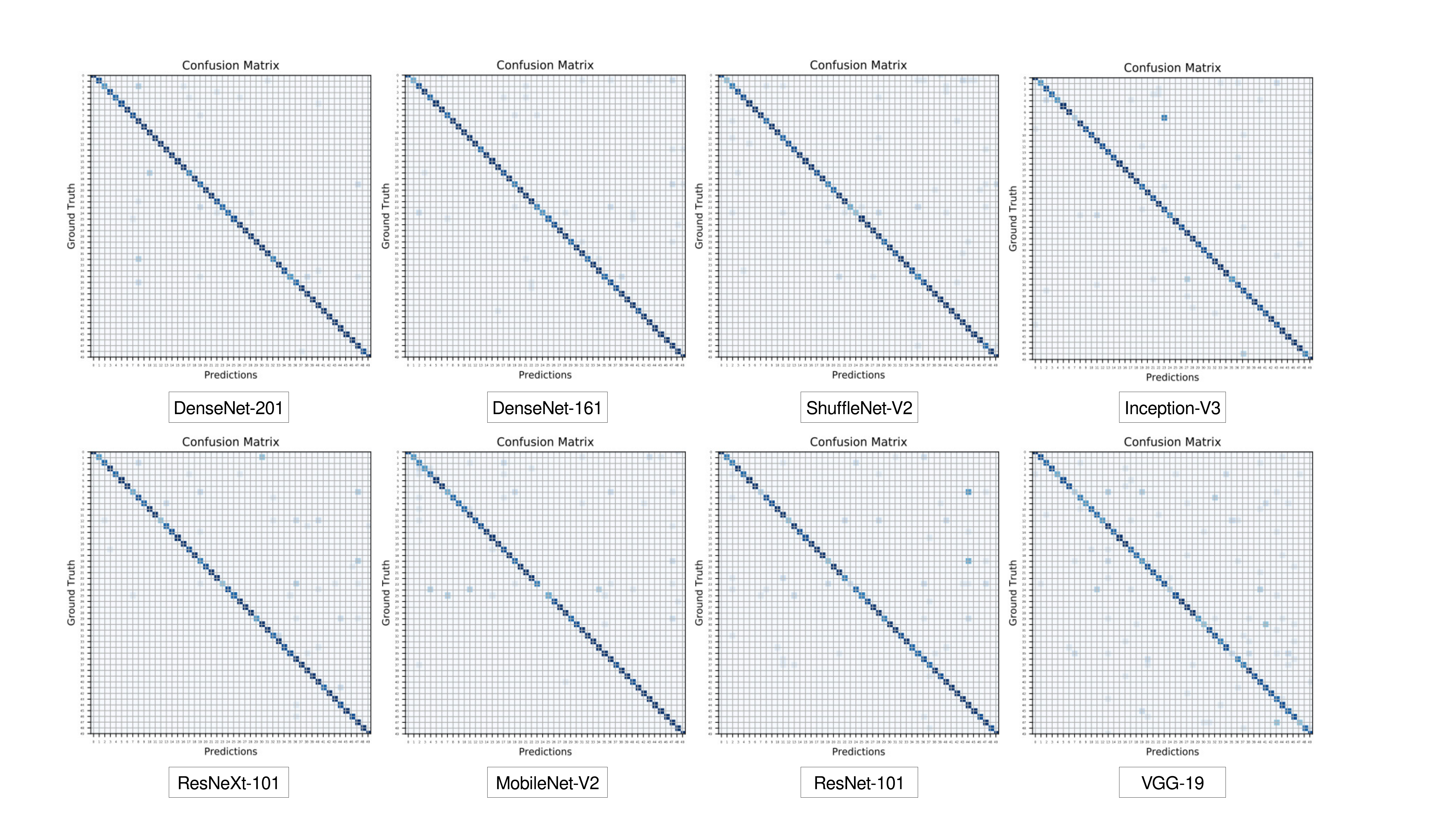}
	\captionsetup[figure]{skip=7pt}
	\captionof{figure}{Confusion matrices for the better performing models (DenseNet-201 \cite{huang2017densely}, DenseNet-161 \cite{huang2017densely}, ShuffleNet-V2 \cite{ma2018shufflenet}, Inception-V3 \cite{szegedy2016rethinking}, ResNeXt-101 \cite{xie2017aggregated}, MobileNet-V2 \cite{sandler2018mobilenetv2}, ResNet-101 \cite{he2016deep} and VGG-19 \cite{simonyan2014very}) trained using our data augmentation techniques.	All the models have been pre-trained on ImageNet. Note that despite the imbalanced training dataset and the difficulty of generalisation, most models are capable of producing accurate and balanced results.}%
	\label{fig:cm}\vspace{-0.3cm}
\end{figure*}

\subsection{Ransomware Classification and Detection}
\label{ssec:lit:ransomware}

Traditionally, malware activities are either detected at the network level \cite{gu2008botminer, cabaj2018software}, system level \cite{bayer2009scalable} or even both \cite{jacob2011jackstraws}. Andronio \cite{andronio2015heldroid} identifies device-locking or encryption activities by finding code paths using static taint analysis along with symbolic execution. Anomalous file system activities have also been used to detect ransomware \cite{kharraz2015cutting}. In another work, abnormal system behaviour is identified based on changes in file type, similarity measurements and entropy \cite{scaife2016cryptolock}.

More recently, machine learning based approaches have become prevalent in detecting and/or classifying ransomware. Sgandurra \etal \cite{sgandurra2016automated} detect and classify ransomware by dynamically analysing the behaviour of applications during the early phases of their installation. In another work \cite{ferrante2017extinguishing}, detection and classification of ransomware is made possible by combining a static detection phase based on the frequency of opcodes prior to installation and a dynamic method which investigates the use of CPU, memory and network as well as call statistics during run-time. Vinayakumar \etal \cite{vinayakumar2017evaluating} explores the use of neural networks with a focus on tuning the hyperparameters and the architecture of a very simple multilayer perceptron to detect and classify ransomware activities.

While the use of various machine learning techniques have led to significant improvements in the field of ransomware detection and classification, these approaches are mostly tailored towards technical users or potential integration into various anti-virus and anti-malware applications. The approach proposed here mainly focuses on classifying ransomware after the system has been infected based on an image of the splash screen or the ransom note casually taken by any layperson.

\subsection{One-Shot Learning}
\label{ssec:lit:one_shot}

Recent advances in modern machine learning techniques have resulted in remarkable strides in various active areas of research, including image classification \cite{simonyan2014very}, semantic scene understanding \cite{atapour2018real, ren2015faster, atapour2019veritatem}, natural language processing \cite{mikolov2013distributed} and graph representations \cite{grover2016node2vec, bonner2018temporal}. However, one of the main requirements of all such approaches is access to a large corpus of data for extensive iterative training, which is often not readily available or intractable to obtain.

This has led to the creation of an entire field of research with a focus on the daunting task of training machine learning algorithms using \emph{one} data point. The seminal work by Fei \etal \cite{fei2006one} popularised the idea of one-shot learning by proposing a variational Bayesian framework for image classification by leveraging previously-learned classes to aid in the classification of unseen ones. Their promising results inspired a slew of researchers to use novel one-shot learning techniques to tackle various other problems and applications. For instance, to address the problem of character classification, Lake \etal model the character drawing process to decompose the image into smaller chucks and a structural explanation is subsequently given for the observed pixels. The same process has been used for speech primitives along with Bayesian inference to identify new words from unknown speakers \cite{lake2014one}.

Siamese neural networks have been used to rank similarity between inputs \cite{koch2015siamese}. This similarity prediction is then utilised to classify not only new data but entirely new classes, by measuring the similarity between the new entries. A memory-augmented neural network is proposed by Santoro \etal \cite{santoro2016one} that learns how to store and retrieve memories to use for each classification task. Vinyals \etal \cite{vinyals2016matching} propose a network that maps a small labelled support set and an unlabelled example to its label, enabling adaptation to new data.

Cheny \etal \cite{cheny2019multi} attempt to learn a mapping between new data samples and \emph{concepts} in a high-dimensional semantic space. The newly mapped concepts are subsequently matched against existing ones and new instance features are synthesised by interpolating among the concepts to facilitate better learning. More similar to our work, Zhao \etal \cite{zhao2019data} directly leverage data augmentation for one-shot learning. In this paper, we also utilise a series of carefully-selected data augmentation techniques to train a classification model based on a single data point per class. Whilst our pipeline is unable to generalise to entirely new classes, we rely on using Bayesian inference to identify previously unseen new classes.

\begin{table}[!t]
	\centering
	\resizebox{\columnwidth}{!}{
		{\tabulinesep=0mm
			\begin{tabu}{@{\extracolsep{5pt}}l c c c c}
				\hline\hline
				\multicolumn{1}{c}{\multirow{2}{*}{Network}} &
				\multicolumn{1}{c}{\multirow{2}{*}{\begin{tabular}{@{}c@{}}Pretrained\\(ImageNet)\end{tabular}}} &  
				\multicolumn{3}{c}{Evaluation Metrics (higher, better)}\T\B \\
				\cline{3-5}
						&& Accuracy	& F\textsubscript{1} Score 	& AUC\T\B \\
				\hline\hline

SqueezeNet\hfill\cite{iandola2016squeezenet}	&\xmark& 0.640 & 0.622 & 0.816\T \\
SqueezeNet\hfill\cite{iandola2016squeezenet}	&\cmark& 0.734 & 0.714 & 0.864 \\

VGG-19\hfill\cite{simonyan2014very}     		&\xmark& 0.670 & 0.661 & 0.832 \\
VGG-19\hfill\cite{simonyan2014very}	           	&\cmark& 0.790 & 0.784 & 0.893 \\

ResNet-101\hfill\cite{he2016deep}				&\xmark& 0.782 & 0.773 & 0.889 \\
ResNet-101\hfill\cite{he2016deep}				&\cmark& 0.876 & 0.872 & 0.937 \\

MobileNet-V2\hfill\cite{sandler2018mobilenetv2}	&\xmark& 0.804 & 0.799 & 0.900 \\
MobileNet-V2\hfill\cite{sandler2018mobilenetv2}	&\cmark& 0.892 & 0.883 & 0.945 \\

ResNeXt-101\hfill\cite{xie2017aggregated}		&\xmark& 0.786 & 0.775 & 0.891 \\
ResNeXt-101\hfill\cite{xie2017aggregated}		&\cmark& 0.898 & 0.896 & 0.948 \\

Inception-V3\hfill\cite{szegedy2016rethinking}	&\xmark& 0.816 & 0.812 & 0.906 \\
Inception-V3\hfill\cite{szegedy2016rethinking}	&\cmark& 0.906 & 0.904 & 0.952 \\

ShuffleNet-V2\hfill\cite{ma2018shufflenet}		&\xmark& 0.774 & 0.764 & 0.885 \\
ShuffleNet-V2\hfill\cite{ma2018shufflenet}		&\cmark& 0.910 & 0.905 & 0.954 \\

DenseNet-161\hfill\cite{huang2017densely}		&\xmark& 0.816 & 0.806 & 0.906 \\
DenseNet-161\hfill\cite{huang2017densely}		&\cmark& 0.928 & 0.926 & 0.963 \\

DenseNet-201\hfill\cite{huang2017densely}		&\xmark& 0.848 & 0.837 & 0.917 \\
DenseNet-201\hfill\cite{huang2017densely}		&\cmark& \textbf{0.936} & \textbf{0.937} & \textbf{0.967}\B \\

\hline \hline
			\end{tabu}
		}
	}
	\captionsetup[table]{skip=7pt}
	\captionof{table}{Results of state-of-the-art classification networks using our data augmentation techniques. Higher resolution images (256 $\times$ 256) are used for training and testing.}
	\label{table:big_networks}\vspace{-0.1cm}
\end{table}

\subsection{Model Uncertainty via Bayesian Approximation}
\label{ssec:lit:bayesian}

In modern applied machine learning, uncertainty is gaining an ever-increasing level of importance, mainly due to the role it can play in detecting and averting adversarial attacks \cite{li2017dropout}, ensuring system safety in critical infrastructure \cite{linda2009neural} and analysing and preventing failure in robotics and navigation applications \cite{kendall2017uncertainties}, among others. Similarly, in our work, uncertainty estimates can be a valuable tool that can ensure new previously-unseen variants of ransomware or completely irrelevant inputs, such as those mistakenly selected by the user, are correctly identified, since explicit handling and treatment is required for these special cases.

A simple and efficient technique widely used in the literature to calculate model uncertainty is Bayesian inference, with dropout \cite{srivastava2014dropout} used as a pragmatic approximation \cite{gal2016dropout}. In a dropout inference approach, the neural network is trained with dropout applied before every weight layer and during testing, the output is obtained by randomly dropping neurons to generate samples from the model distributions. Gal \etal \cite{gal2016dropout} demonstrate that the use of dropout inference is mathematically equivalent to the probabilistic deep Gaussian process approximation \cite{damianou2013deep}, with the approach effectively minimising the Kullback-Leibler divergence between the model distribution and the posterior of a deep Gaussian process, marginalised over its finite rank covariance function parameters \cite{gal2016dropout}.

While the use of such an approach \cite{gal2016dropout} can yield a reasonable estimate of model uncertainty (as demonstrated in Section \ref{ssec:experiments:uncertainty}), to obtain better-calibrated uncertainty that fits the nature of the data at hand, the dropout rate at each layer must be adapted to the data as a variational parameter. This is often accomplished using an extensive grid-search \cite{gal2017concrete} which is computationally-intensive, time-consuming, and completely intractable for certain tasks, which points to the importance of an adaptive dropout rate in a variational framework.

\begin{table}[!t]
	\centering
	\resizebox{\columnwidth}{!}{
		{\tabulinesep=0mm
			\begin{tabu}{@{\extracolsep{5pt}}c c c c c @{}}
				\hline\hline
				\multicolumn{1}{c}{\multirow{2}{*}{Network}} &
				\multicolumn{1}{c}{\multirow{2}{*}{\# Parameters}} &  
				\multicolumn{3}{c}{Evaluation Metrics (higher, better)}\T\B \\
				\cline{3-5}
						&& Accuracy	& F\textsubscript{1} Score 	& AUC\T\B \\
				\hline\hline

Inception-V3\hfill\cite{szegedy2016rethinking}	& 25,214,714 & 0.626 & 0.591 & 0.809\T \\

ShuffleNet-V2\hfill\cite{ma2018shufflenet}		& 1,304,854  & 0.628 & 0.604 & 0.810 \\

VGG-19\hfill\cite{simonyan2014very}	        	& 139,786,098& 0.630 & 0.609 & 0.811 \\

SqueezeNet\hfill\cite{iandola2016squeezenet}	& \textbf{748,146}	 & 0.634 & 0.613 & 0.813 \\

ResNet-101\hfill\cite{he2016deep}				& 42,602,610 & 0.664 & 0.642 & 0.829 \\

MobileNet-V2\hfill\cite{sandler2018mobilenetv2}	& 2,287,922  & 0.666 & 0.648 & 0.830 \\

ResNeXt-101\hfill\cite{xie2017aggregated}		& 86,844,786 & 0.674 & 0.659 & 0.834 \\

DenseNet-201\hfill\cite{huang2017densely}		& 18,188,978 & 0.720 & 0.704 & 0.857 \\

DenseNet-161\hfill\cite{huang2017densely}		& 26,582,450 & \textbf{0.744} & \textbf{0.734} & \textbf{0.870}\B \\

\hline

Custom Network	& 1,875,666  & 0.716 & 0.703 & 0.855\T\B \\

\hline\hline
			\end{tabu}
		}
	}
	\captionsetup[table]{skip=7pt}
	\captionof{table}{Results of state-of-the-art classification architectures and our custom-made light-weight network. Lower resolution images (128 $\times$ 128) are used to reduce the number of parameters and increase the rate of convergence.}
	\label{table:small_networks}\vspace{-0.2cm}
\end{table}

Kingma \etal \cite{kingma2015variational} thus propose variational dropout, which attempts to model Bayesian inference using a posterior factorised over individual network weights $w_i \in \bm{W}, q(w)=\mathcal{N}(\theta, \alpha\theta^2)$ for all individual mean parameters $\theta_i \in \bm{\theta}$. The prior factorises similarly and is explicitly selected so the Kullback-Leibler divergence between the model distribution and the posterior $q(\bm{W})$ is independent of the mean parameters $\bm{\theta}$. Additionally, Kingma \etal \cite{kingma2015variational} claim that their reparametrisation technique maps uncertainty about the weights of the model into independent local noise. Subsequently, an extension to the conventional Gaussian multiplicative dropout \cite{srivastava2014dropout} is proposed that allows for the dropout rate to be learned as a parameter. However, more recent studies \cite{hron2017variational, hron2018variational} have demonstrated that the log-uniform prior used for variational dropout \cite{kingma2015variational} may not lead to a proper posterior, which means variational dropout is a non-Bayesian sparsification approach and the uncertainty estimated based on $q(\bm{W})$ may not follow the usual Bayesian interpretation.

Conversely, Gal \etal \cite{gal2017concrete} resolve the issue of the improper prior and posterior and propose the use of learnable dropout rate parameters optimised towards obtaining better uncertainty rather than maximising model performance. By introducing a dropout regularisation term, which only depends on the dropout rate, the approach ensures that the posterior approximated by the dropout itself does not deviate too far from the model distribution. In this paper, we make use of all three approaches \cite{gal2016dropout, kingma2015variational, gal2017concrete} to obtain uncertainty and assess the performance and efficacy of each using our data.


\section{Approach}
\label{sec:approach}

The primary objective of this work is to investigate the possibility of classifying the variant of ransomware a system is infected with solely based on an image of the splash screen or the ransom note captured from a computer screen (or mobile device) using a consumer camera. This is accomplished by training a classifier on a single original image of the splash screen of each ransomware. In the following, we will outline the details of the our dataset, data augmentation techniques and the different networks used to carry out the classification.

\subsection{Training Dataset}
\label{ssec:approach:dataset}

To explore the potentials of our ransomware classification pipeline, we train our model on a dataset of splash screens and ransom notes of 50 different variants of ransomware. A single image of a splash screen variant is available for each of the ransomware classes available in our dataset. However, certain ransomware classes are associated with more than one splash screen (\ie certain classes contain more than one training image but those images depict different splash screens associated with the same class), which significantly adds to the difficulty of the problem as this leads to a training data imbalance and can lead to training instability.

\begin{table}[!t]
	\centering
	\resizebox{\columnwidth}{!}{
		{\tabulinesep=0mm
			\begin{tabu}{@{\extracolsep{5pt}}c c c c @{}}
				\hline\hline
				\multicolumn{1}{c}{\multirow{2}{*}{Augmentation Method}} & 
				\multicolumn{3}{c}{Evaluation Metrics (higher, better)}\T\B \\
				\cline{2-4}
						& Accuracy	& F\textsubscript{1} Score 	& AUC\T\B \\
				\hline\hline
None					& 0.252 			& 0.258 					& 0.618\T \\
Contrast				& 0.386 			& 0.379 					& 0.687 \\
Rotation				& 0.440 			& 0.414 					& 0.714 \\
Brightness				& 0.404 			& 0.402 					& 0.696 \\
Perspective				& 0.524 			& 0.500 					& 0.757 \\
Motion Blur				& 0.338 			& 0.348 					& 0.662 \\
Defocus Blur			& 0.324 			& 0.324 					& 0.655 \\
Gaussian Blur			& 0.312 			& 0.289 					& 0.649 \\
Random Noise			& 0.344 			& 0.343 					& 0.665 \\
Random Occlusion 		& 0.344 			& 0.339 					& 0.665 \\
Colour Perturbations	& 0.330 			& 0.325 					& 0.658\B \\
\hline
All Augmentations		& \textbf{0.716} 	& \textbf{0.703} 	& \textbf{0.855}\T\B \\
\hline
			\end{tabu}
		}
	}
	\captionsetup[table]{skip=7pt}
	\captionof{table}{Numerical results demonstrating the importance of the augmentation techniques (Section \ref{ssec:approach:augmentation}) used for training.}
	\label{table:augmentations}\vspace{-0.5cm}
\end{table}

To test the performance of the approach, a balanced test set of 500 images (10 images per class) is created by casually taking screenshots of the ransomware images using two different types of camera phones (Apple and Android) from 6 different computer screens (with varying specifications, \eg size, resolution, aspect ratio, panel type, screen coating and colour depth). We call this the \emph{positive} test dataset since all the images within this dataset need to be \emph{positively} identified as ransomware and any model trained using our dataset should be certain about the predictions it makes with respect to the ransomware variants it has already observed. 

An additional set of 50 unrelated and/or non-ransomware images are captured from the same computer screens (under the same conditions as our positive test images) to evaluate the uncertainty estimates acquired using our Bayesian networks. We refer to this portion of our dataset as the \emph{negative} test dataset, as any model trained on our dataset should be very \emph{uncertain} about this data since these screenshot images are not of, and therefore should not be classified as, any ransomware known to the model. Examples of the training and positive and negative test images are shown in Figure \ref{fig:taste_data}. Note that some of the images in our negative test set (Figure \ref{fig:taste_data} -- \emph{bottom}) are very similar in appearance to what a ransomware splash screen could look like. This has been purposefully designed so the uncertainty values estimated by the model can be more rigorously assessed.

Using our carefully designed augmentation techniques, we train the models on our training dataset of 66 images in 50 classes. In the following, we will briefly outline the details of our data augmentation techniques.
\begin{table*}[!t]
	\centering
	\resizebox{0.9\linewidth}{!}{
		{\tabulinesep=0mm
			\begin{tabu}{@{\extracolsep{5pt}}c c c c c c c c @{}}
				\hline\hline
				\multicolumn{1}{c}{\multirow{2}{*}{Augmentation}} & 
				\multicolumn{3}{c}{Evaluation Metrics (higher, better)} &
				\multicolumn{1}{c}{\multirow{2}{*}{Augmentation}} & 
				\multicolumn{3}{c}{Evaluation Metrics (higher, better)}\T\B \\
				\cline{2-4} \cline{6-8}
						& Accuracy	& F\textsubscript{1} Score 	& AUC & & Accuracy	& F\textsubscript{1} Score 	& AUC\T\B \\
				\hline\hline
P/R/B/C/N/O/M/CP/D/G	& \textbf{0.716} 	& \textbf{0.703} 	& \textbf{0.855} &  P/R/B/C/N	& 0.616 & 0.609 & 0.782\T \\
P/R/B/C/N/O/M/CP/D		& 0.690 			& 0.681 					& 0.842  &  P/R/B/C	& 0.606 & 0.592 & 0.776 \\
P/R/B/C/N/O/M/CP		& 0.674 			& 0.658 					& 0.821  &  P/R/B	& 0.592 & 0.580 & 0.771 \\
P/R/B/C/N/O/M			& 0.648 			& 0.632 					& 0.805  &  P/R		& 0.586 & 0.569 & 0.762 \\
P/R/B/C/N/O				& 0.634 			& 0.628 					& 0.797  &  P		& 0.524 & 0.500 & 0.757\B \\
\hline \hline
			\end{tabu}
		}
	}
	\captionsetup[table]{skip=7pt}
	\captionof{table}{Evaluating the performance of the combined augmentation techniques. \textbf{C:} Contrast; \textbf{R:} Rotation; \textbf{B:} Brightness; \textbf{P:} Perspective; \textbf{M:} Motion blur; \textbf{D:} Defocus blur; \textbf{G:} Gaussian blur; \textbf{N:} Noise; \textbf{O:} Occlusion; \textbf{CP:} Colour Perturbation.}
	\label{table:augmentations_combined}\vspace{-0.4cm}
\end{table*}
\subsection{Data Augmentation}
\label{ssec:approach:augmentation}

During training, the network can only see the single image available for each splash screen variant. This lack of training data can significantly hinder generalisation as the model would simply overfit to the training distribution or memorise the few training images it has seen. This means a model trained on our training dataset without any modification or augmentation would be incapable of classifying images captured under test conditions from a computer screen (Section \ref{ssec:experiments:ablation}).

To prevent this, a carefully designed and tuned set of augmentation techniques is applied to the training images on the fly to simulate the test conditions (images casually captured from a computer screen). The hyper-parameters associated with these augmentation techniques (\eg thresholds, intensity) are determined using exhaustive grid-searches which are excluded here. Each of the following augmentation techniques is randomly applied (both in terms of application and severity):
\begin{enumerate*}[label=(\arabic*)]
	\item \emph{rotation:} randomly rotating the image with the angle of rotation in the range [-90\degree,90\degree],
	\item \emph{contrast:} randomly changing the image contrast by up to a factor of 2,
	\item \emph{brightness:} randomly changing the brightness by up to a factor of 3,
	\item \emph{occlusion:} to primarily simulate distractors such as screen glare and reflection mostly in glossy screens (up to a quarter of the image size occluded with random elliptical shapes of randomly selected bright colours),
	\item \emph{Gaussian blur:} with a radius of up to 5,
	\item \emph{motion blur:} simulating blurring effects caused by the movement of the camera during image capture (up to a movement length of 9 pixels -- see Figure \ref{fig:augs} - bottom),
	\item \emph{defocus blur:} simulating the camera being out of focus which is a common occurrence when a computer screen is being photographed (up to a kernel size of 9 -- see Figure \ref{fig:augs} - top),
	\item \emph{noise:} Gaussian noise up to a level of 0.2,
	\item \emph{colour perturbations:} randomising hue by a maximum of 5\% and saturating colours by a factor of up to 2, and
	\item \emph{perspective:} by up to 50\% over each axis to simulate the varying camera angles when a screen is being photographed.
\end{enumerate*}

By using random combinations of all the different augmentation methods applied to our training set with varying probabilities, very high levels of accuracy can be achieved (see Section \ref{sec:experiments}). In the following section, we will focus on the details of the classification models and the network architectures that take advantage of these data augmentation techniques used within our approach to classify ransomware based on our training dataset.

\subsection{Classification Model}
\label{ssec:approach:model}

A very effective way of solving the problem of ransomware classification is to use to the augmentation methods outlined in Section \ref{ssec:approach:augmentation} along with any of the many optimised classification networks in the literature \cite{simonyan2014very, he2016deep, iandola2016squeezenet, huang2017densely, szegedy2016rethinking, ma2018shufflenet, sandler2018mobilenetv2, xie2017aggregated}. Most of these networks are capable of yielding very high-accuracy results, especially when taking advantage of the boosted features that can be obtained by pre-training the network on large datasets such as ImageNet (Table \ref{table:big_networks}). However, it is important to note that despite the recent introduction of more efficient light-weight networks \cite{iandola2016squeezenet, ma2018shufflenet, sandler2018mobilenetv2}, the majority of the state-of-the-art classification models make use of very deep architectures and contain an extremely large number of parameters (Table \ref{table:small_networks}).

An important part of this work is to enable an accurate measurement of model uncertainty via Bayesian approximation, and as explained in Section \ref{ssec:lit:bayesian}, this can be accomplished with a reasonable degree of mathematical accuracy by applying a dropout layer before every weight layer within the model. This can be highly problematic for very deep networks \cite{simonyan2014very, szegedy2016rethinking, he2016deep} since the large number of dropout layers in such networks would make convergence intractable. While simply reducing the number of dropout layers in a very deep network can help with the convergence problem \cite{gal2016dropout}, it comes at a cost of the precision of the uncertainty values since it would not be possible to accurately calibrate the uncertainty estimation process if some layers contain neurons that cannot be dropped.

\begin{figure}[t!]
	\centering
	\includegraphics[width=0.99\linewidth]{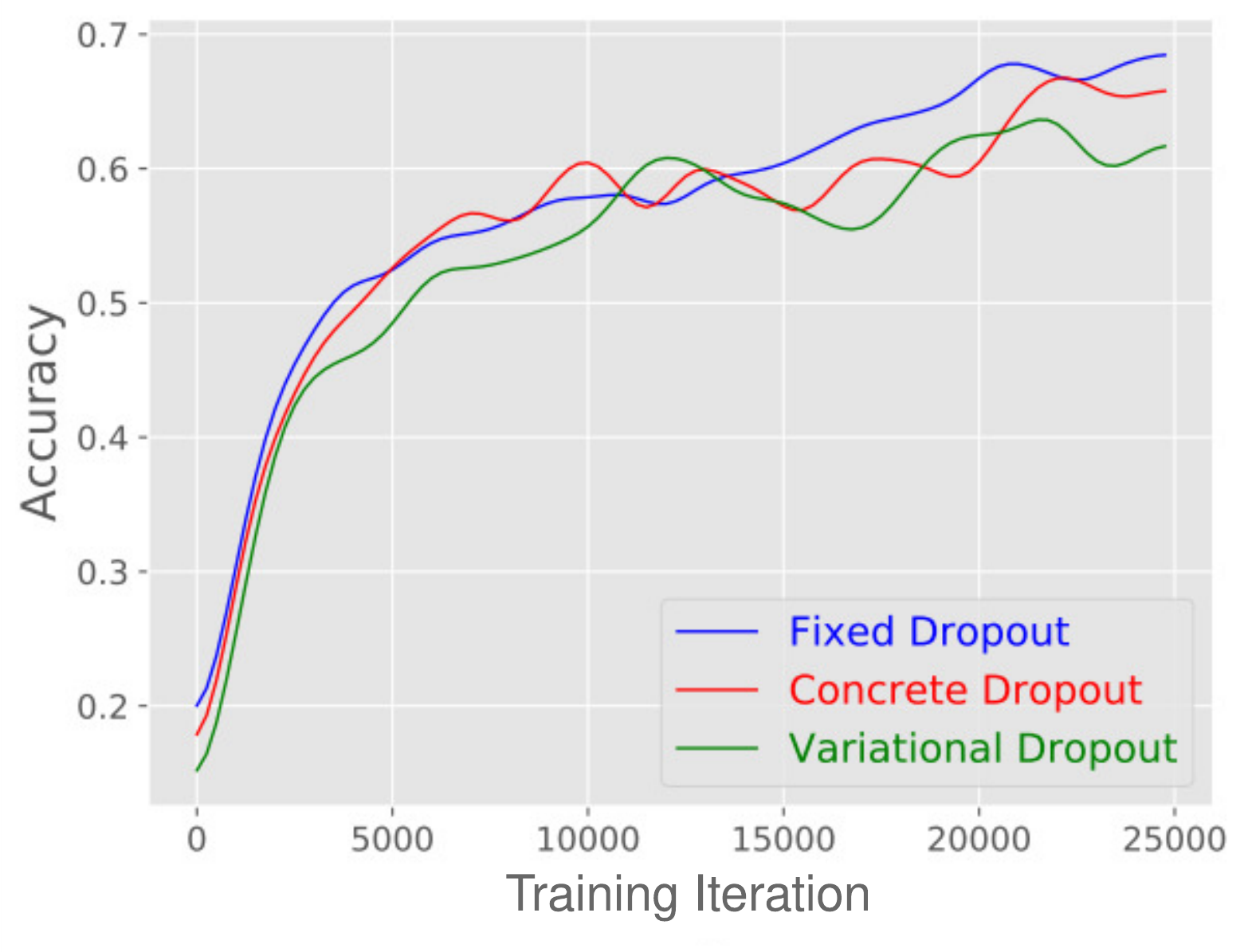}
	\captionsetup[figure]{skip=7pt}
	\captionof{figure}{Test accuracy of our custom network with fixed dropout \cite{gal2016dropout}, concrete dropout \cite{gal2017concrete} and variational dropout \cite{kingma2015variational} layers as the models are trained for 25,000 iterations.}%
	\label{fig:acc}\vspace{-0.5cm}
\end{figure}

To remedy this issue and for the sake of experimental consistency, we propose a simplified custom architecture, seen in Figure \ref{fig:arch}. This light-weight network takes an image of size $128 \times 128$ as its input and after six convolutional layers and three max-pooling operations produces a feature vector of 4096 dimensions. This is subsequently passed into a fully-connected layer which classifies the input into one of 50 classes. Training is accomplished via a cross entropy loss function. No normalization is performed in the network. To approximate Bayesian inference, a dropout layer can be placed after every weight layer in the network. Figure \ref{fig:arch} shows an outline of our custom network architecture, with the dropout layers optionally used to approximate Bayesian inference.

\begin{figure*}[t!]
	\centering
	\includegraphics[width=0.99\linewidth]{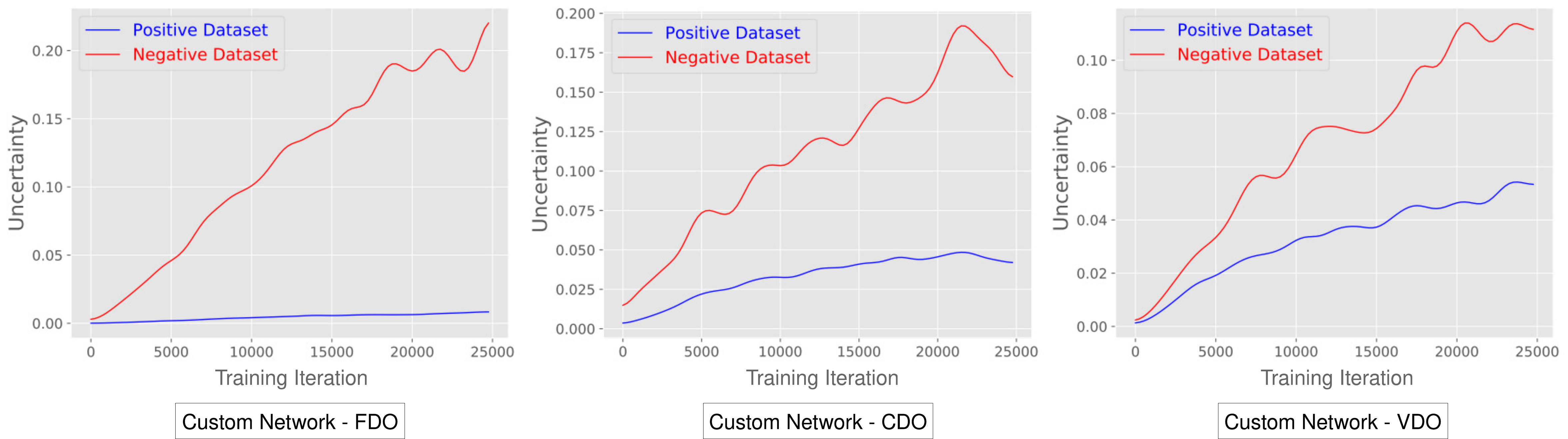}
	\captionsetup[figure]{skip=7pt}
	\captionof{figure}{Comparing the uncertainty values as our custom network is trained with fixed dropout \cite{gal2016dropout} (FDO), concrete dropout \cite{gal2017concrete} (CDO) and variational dropout \cite{kingma2015variational} (VDO) layers. All models demonstrate higher levels of uncertainty on the negative test dataset (\emph{red}) than on the positive test images (\emph{blue}).}%
	\label{fig:uncer_plots}\vspace{-0.5cm}
\end{figure*}

We utilise the Bayesian dropout techniques \cite{gal2016dropout, kingma2015variational, gal2017concrete} to calculate model uncertainty via Monte Carlo sampling. After $N$ stochastic forward passes of the same input $\bm{\mathrm{X}}$ (images) through the network to produce the output $\bm{\mathrm{Y}}$ (class labels), the predictive mean of the model is as follows:
\begin{equation}
	\mathbb{E}(\bm{\mathrm{Y}}) = \frac{1}{N} \sum_{n=1}^{N} \bm{\mathrm{Y}}'_n .
	\label{eq:pred_mean}\vspace{-0.1cm}
\end{equation}
The predictive uncertainty is thus obtained as follows:
\begin{equation}
	\mathrm{Var}(\bm{\mathrm{Y}}) = \frac{1}{N} \sum_{n=1}^{N} {\bm{\mathrm{Y}}'_n}^T \bm{\mathrm{Y}}'_n - {\mathbb{E}(\bm{\mathrm{Y}})}^T \mathbb{E}(\bm{\mathrm{Y}}) .
	\label{eq:pred_var}\vspace{-0.1cm}
\end{equation}

The dropout rate can be set as a fixed hyper-parameter tuned to the data via intensive grid-searches (0.05 in our case for all six dropout layers in the network) or learned as model parameters \cite{kingma2015variational, gal2017concrete}. In Section \ref{ssec:experiments:uncertainty}, we experiment with all these variations of Bayesian approximation through dropout to enable further insight into the functionality of our model and uncertainty measurements in general.

\subsection{Implementation Details}
\label{ssec:approach:implementation-details}

The image data in our training and test sets are all of different resolutions but for the sake of consistency, they are all cropped to a square with the length equal to the smaller dimension of the image (random cropping for training images and centre cropping for test images) and resized to an image of dimensions $128 \times 128$ for our custom network architecture or $256 \times 256$ to achieve higher accuracy results using deeper convolutional networks. The non-linearity module used in our custom architecture is leaky ReLU ($slope=0.2$). The training data imbalance issue is handled by weighting the inputs in the loss function according to the frequency of their class within the overall dataset. All models are trained to 100,000 steps. The implementation is done in \emph{PyTorch} \cite{pytorch}, with Adam \cite{kingma2014adam} empirically providing the best optimization ($\beta_{1} = 0.5$, $\beta_{2} = 0.999$, $\alpha = 0.0002$).

\section{Experimental Results}
\label{sec:experiments}

In this section, we evaluate our work using extensive experimental analysis. The results of various state-of-the-art classification approaches are evaluated and using ablation studies, we demonstrate the importance of our data augmentation approaches. Additionally, using our positive and negative test datasets, we investigate the effectiveness of model uncertainty values obtained through Bayesian approximation via dropout.  

\subsection{State-of-the-Art Classification}
\label{ssec:experiments:classification}

To achieve the highest possible levels of accuracy, we train various state-of-the-art image classification networks \cite{simonyan2014very, he2016deep, iandola2016squeezenet, huang2017densely, szegedy2016rethinking, ma2018shufflenet, sandler2018mobilenetv2, xie2017aggregated}. With relatively high-resolution images (256 $\times$ 256) used as inputs, accuracy levels of up to 93.6\% can be achieved using our full augmentation protocol and a DenseNet-201 network \cite{huang2017densely} pre-trained on ImageNet.

Table \ref{table:big_networks} contains the numerical results obtained from different architectures across various metrics with inputs of size 256 $\times$ 256. As seen in Table \ref{table:big_networks}, the representation learning encapsulated within the model resulting from the features obtained by pre-training the network on ImageNet is an invaluable asset and can lead to performance boosts of up to 14\% for some of the networks.

As indicated by the high F\textsubscript{1} score, despite the uneven class distribution in our training dataset, using our class balancing efforts (Section \ref{ssec:approach:implementation-details}), most networks are capable of learning about all the classes in an evenly distributed manner. The high AUC (Area Under the Curve) values also demonstrate the great leaning capabilities of the models which are able to easily distinguish between the classes with little confusion. The confusion matrices for some of the models \cite{simonyan2014very, he2016deep, huang2017densely, szegedy2016rethinking, ma2018shufflenet, sandler2018mobilenetv2, xie2017aggregated} shown in Figure \ref{fig:cm} further confirm these findings and point to the strong feature learning capabilities of the models.

\begin{figure*}[t!]
	\centering
	\includegraphics[width=0.99\linewidth]{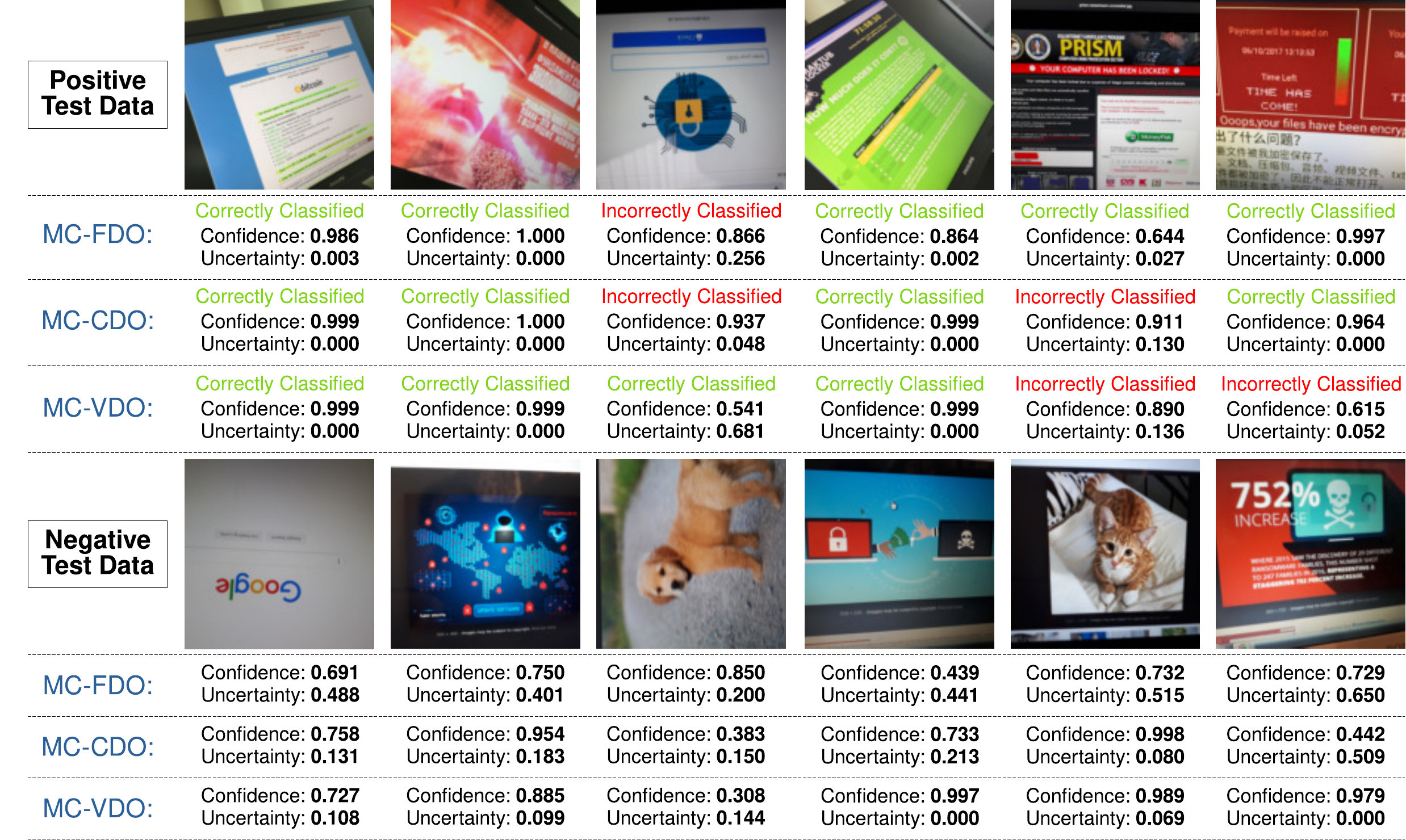}
	\captionsetup[figure]{skip=7pt}
	\captionof{figure}{Examples demonstrating the effectiveness of model uncertainty, using positive test data (screenshots of images from the training set -- \emph{top}) and negative test data (unrelated images that should not be classified -- \emph{bottom}). \textbf{MC:} Monte Carlo sampling; \textbf{FDO:} Fixed dropout \cite{gal2016dropout}; \textbf{CDO:} Concrete dropout \cite{gal2017concrete}; \textbf{VDO:} Variational dropout \cite{kingma2015variational}.}%
	\label{fig:uncertainty}\vspace{-0.3cm}
\end{figure*}

An important aspect of our work, however, is training and inference efficiency. Fast convergence during training can be intractable in very deep models when dropout is utilised as Bayesian approximation to obtain model uncertainty. Since our approach is meant to specifically accommodate lay users through a web server, a light-weight model that can be used for efficient deployment is very important to reduce the chance of high load and hence denial of service.

To address the issues of efficiency and convergence rate and to guarantee better experimental consistency, we evaluate our custom architecture network that takes smaller (128 $\times$ 128) images as its input compared against state-of-the-art deep and light-weight networks commonly used within the literature, when receiving the same small (128 $\times$ 128) images as their input. As seen in Table \ref{table:small_networks}, our simpler network outperforms most deeper and light-weight networks \cite{simonyan2014very, he2016deep, iandola2016squeezenet, szegedy2016rethinking, ma2018shufflenet, sandler2018mobilenetv2, xie2017aggregated} while remaining competitive with the rest \cite{huang2017densely}. The superior performance of our simple architecture is mainly due to the fact that the number of its layers and parameters are carefully tuned to the dataset (using preliminary architecture searches, which have been excluded for brevity).

\subsection{Ablation Studies}
\label{ssec:experiments:ablation}

One of our primary contributions is the ability to train an accurate ransomware screenshot classifier using a single training image for each variant of splash screen or ransom note. This is achieved using ten carefully-designed augmentation techniques (Section \ref{ssec:approach:augmentation}), the combination of which will result in the simulation of a screenshot of a ransomware splash screen captured using a consumer-grade camera. Consequently, a substantial part of our experimental setup has been to demonstrate the importance of each of these augmentation techniques to ensure that they positively contribute to the improved performance of the model. To accomplish this, we train our custom network (with no dropout) using individual augmentation techniques to measure their effects on the results. Table \ref{table:augmentations} contains the results of our custom network when trained on individual augmentation methods.

As expected, not using any augmentation leads to a poor performance from the model, while significantly better results can be achieved when all the augmentation methods are combined. We also experimented with random combinations of the techniques to empirically investigate any incompatibility, but found that all augmentation techniques used here contribute to the improvement of the results, as seen in Table \ref{table:augmentations_combined}.

\begin{table*}[!t]
	\centering
	\resizebox{\textwidth}{!}{
		{\tabulinesep=0mm
			\begin{tabu}{@{\extracolsep{7pt}}l c c c c c c c@{}}
				\hline\hline
				\multicolumn{1}{l}{\multirow{2}{*}{Approach}} &
				\multicolumn{1}{c}{\multirow{2}{*}{}} &  
				\multicolumn{1}{c}{\multirow{2}{*}{Test Data}} &  
				\multicolumn{3}{c}{Evaluation Metrics (higher, better)} &
				\multicolumn{2}{c}{Uncertainty and Confidence}\T\B \\
				\cline{4-6} \cline{7-8}
					&	&  &Accuracy	& F\textsubscript{1} Score 	& AUC & Model Uncertainty & Mean Confidence\T\B \\
				\hline\hline
\multirow{2}{*}{Fixed Dropout} & \multirow{2}{*}{\cite{gal2016dropout}}	& Positive & 0.708 & 0.7011	& 0.8429 & 0.015 & 0.85 $\pm$ 0.21\T\B \\
										& & Negative & -- & -- & -- & 0.330 & 0.66 $\pm$ 0.25\B \\
\hline

\multirow{2}{*}{Concrete Dropout} & \multirow{2}{*}{\cite{gal2017concrete}}	& Positive & 0.698 & 0.6771	& 0.8459 & 0.067 & 0.87 $\pm$ 0.19\T\B \\
										& & Negative & -- & -- & -- & 0.218 & 0.72 $\pm$ 0.29\B \\
\hline

\multirow{2}{*}{Variational Dropout} & \multirow{2}{*}{\cite{kingma2015variational}}	& Positive & 0.6821 & 0.6593	& 0.8378 & 0.084 & 0.86 $\pm$ 0.22\T\B \\
										& & Negative & -- & -- & -- & 0.175 & 0.71 $\pm$ 0.23\B \\
\hline\hline
			\end{tabu}
		}
	}
	\captionsetup[table]{skip=7pt}
	\captionof{table}{Numerical results of different Bayesian approximation methods \cite{gal2016dropout, gal2017concrete, kingma2015variational} to obtain model uncertainty. As expected the models have low uncertainty for the positive test data (screenshots of ransom notes in the training set) and high uncertainty for negative test images (unrelated images and new ransomware variants)}
	\label{table:uncertainty}\vspace{-0.5cm}
\end{table*}

As seen in Tables \ref{table:augmentations} and \ref{table:augmentations_combined}, perspective and rotation have the greatest influence over the results. In our experiments with additional augmentation techniques, we found that horizontally flipping the images results in worse model performance since the test set does not contain any mirror images, as modern consumer cameras do not produce mirrored outputs. We also interestingly found that adding vertical flipping to the mix of our augmentation techniques had no impact on the results as the effects of this augmentation methods can be achieved through rotation. As a result, image flipping was removed from the list of augmentation techniques used in our approach.

\subsection{Model Uncertainty}
\label{ssec:experiments:uncertainty}

Another important component of this work is the ability of the model to calculate uncertainty, therefore enabling the identification of unrelated input images (\eg non-ransomware inputs and new previously-unseen ransomware images). Our custom network (Figure \ref{fig:arch}) is consequently trained with the three different dropout modules \cite{gal2016dropout, gal2017concrete, kingma2015variational} used for Bayesian approximation. Dropout layers are kept in place during inference and uncertainty is calculated as per Eqn. \ref{eq:pred_var} via Monte Carlo sampling of the network weights. Recent work \cite{hron2017variational, hron2018variational} argue that the use of variational dropout \cite{kingma2015variational} does not lead to proper Bayesian behaviour and can result in overfitting. This notion is somewhat confirmed by our experiments. A seen in Figure \ref{fig:acc}, our network trained with variational dropout is prone to overfitting and produces lower test accuracy levels.

Moreover, by calculating model uncertainty when the model is evaluated using our positive and negative test data, we can assess the effectiveness of our uncertainty values. One would expect the model to be very uncertain when negative test images (unrelated images) are passed in as inputs, while the uncertainty values should be smaller when positive test data (ransomware screenshots) are seen by the network. As seen in Figure \ref{fig:uncer_plots}, our experiments point to the same conclusions with uncertainty values being significantly higher in the presence of negative data. Interestingly, as seen in Figure \ref{fig:uncer_plots}, a fixed dropout rate (FDO) \cite{gal2016dropout} produces cleaner and more accurate uncertainty values despite the intensive computation required to determine the dropout rate (0.05 for all layers in our case).

Figure \ref{fig:uncertainty} shows the confidence and uncertainty values obtained for a small number of randomly-selected examples from our positive and negative test datasets. As expected, confidence values (softmax outputs) are essentially meaningless and contain very little information about how much the network actually knows about the image, while uncertainty values are a better indicator of whether the network has sufficient knowledge of the input image or not. For our best-performing model (fixed dropout), an uncertainty value of 0.12 seems to be a reasonable estimated threshold, beyond which the predictions of the model are not credible. Similar conclusions can be drawn from Table \ref{table:uncertainty}, which contains the numerical results of the Bayesian approximation methods \cite{gal2016dropout, gal2017concrete, kingma2015variational} applied to positive and negative test data. As seen in Table \ref{table:uncertainty}, the mean uncertainty values are an order of magnitude higher for the negative test images than they are for the positive images, and the confidence values have such a high standard deviation that their use to measure how much the model knows about the input it has received can lead to very misleading results.

\section{Discussions and Future Work}
\label{sec:discussions}

As discussed in Section \ref{ssec:experiments:classification}, we are able to achieve high accuracy results using our augmentation techniques and deep convolutional neural networks such as DenseNet \cite{huang2017densely}. However, since another important component of our work, model uncertainty, relies on introducing a dropout layer after every weight layer within the model, convergence for very deep models such as DenseNet \cite{huang2017densely} would be almost impossible, which is why we opt for using our own simplified network architecture. 

While this can sufficiently meet the requirements of our application through a possible two-stage solution (the light-weight network measures the uncertainty of the model with respect to the input and if the value is low and special handling is not required, the deep network can be subsequently used to conduct the actual classification), future work can involve the use of Bayesian modules within each layer \cite{tran2018bayesian} or a Bayesian last layer in the network \cite{Weber2018optimizing}, thus enabling the optimisation of much deeper networks with plausible uncertainty calculation capabilities. Additionally, if the parameters of the augmentation techniques could be learned during training instead of being laboriously tuned through extensive grid searches, the resulting efficient and stable training procedure can lead to superior model performance.

\section*{Conclusion}

In this work, we explore the possibility of performing the task of ransomware classification based on a simple screenshot of the splash screen or ransom note captured using a consumer camera found in any modern mobile phone. To make this possible, we create a sample dataset with only a single image available for every variant of ransomware splash screen. Instead of creating a large corpus of ransomware screenshot images for training, we opt for simulating the conditions that lead to the appearance of a screenshot image through carefully designed data augmentation techniques, resulting in a very simple one-shot learning procedure. Additionally, we employ various Bayesian approximation approaches \cite{gal2016dropout, gal2017concrete, kingma2015variational} to obtain model uncertainty. Using uncertainty values, we are then able to identify special input cases such as unrelated non-ransomware images and new previously-unseen ransomware variants that our trained models are able or expected to classify. These particular input cases can be set aside for special handling. Using extensive experimental evaluation, we have demonstrated that test accuracy levels of up to 93.6\% can be achieved using our full augmentation protocol and a deep network such as DenseNet \cite{huang2017densely}. Assessments using our negative test dataset (images unknown to the model) also indicate that our custom architecture trained with \cite{gal2016dropout, gal2017concrete, kingma2015variational} is capable of accurately estimating uncertainty values.

\section*{Acknowledgement}

We would like to thank the Engineering and Physical Sciences Research Council (EPSRC) for funding this research project. This work in part made use of the Rocket High Performance Computing service at Newcastle University.


\bibliographystyle{IEEEtran}
\bibliography{paper_ref}

\end{document}